# Multi-Objective Evolutionary Algorithms platform with support for flexible hybridization tools

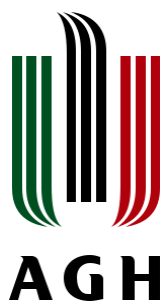

mgr inż. Michał Idzik

Wydział Informatyki, Elektroniki i Komunikacji

Akademia Górniczo-Hutnicza

# Contents

# 1

# Introduction

Picking the best solution in decision-making processes very often is a complex and demanding task. Real-world problems usually consist of multiple, often conflicting objectives or criteria that need to be considered. In such processes de-cision maker has to choose the most satisfying solution from a set of equally good, optimal ones. The main problem here is how to find this optimal set (represented as Pareto set), assuming we have knowledge about our domain and we know how to evaluate each criterion on proposed solution data.

If we express our objectives in mathematical terms, as functions mapping solution vector into quality indicator, we can define Multi-objective Optimization Problem (MOP) and apply one of common methods. However, the most suitable approaches producing whole Pareto set are *a posteriori* methods. Multi-objective Optimization Evolutionary Algorithms (MOEAs) are very popular *a posteriori* class representants. MOEAs are widely used with satisfactory results, though they are also frequently criticized due to their high computational complexity. What is more, they exhibit several difficulties mainly caused by the high dimensionality of the search space and the multimodality of the objective functions. These types of obstacles, as well as high cost of accurate single fitness evaluation, are common features of real-world multi-objective and inverse problems [5, 8].

In order to efficiently explore the search space in such cases right from the beginning of computations and to find all the connected components of the Pareto set, we have introduced the maturing Hierarchical Genetic Strategy for Multi-objective Optimization (MO-mHGS) [7]. This strategy improves and extends the concept of MO-HGS [4], a method combining the multi-deme hierarchical strategy HGS [9] with a genetic engine driven by MOEA. Our recent research has shown MO-mHGS explores the search space effectively and can be coupled with a variety of internal, single-deme algorithms. It quickly approaches the optimal Pareto front, often achieving closer proximity than its competitors. Its main advantage occurs in the early stage of computations, where its fast convergence with a small budget can be even further exploited by incorporating a local gradient-based algorithm, and in problems for which time of calculating a single fitness evaluation depends progressively on the evaluation accuracy. Moreover, tree-based structure composed of multiple independent nodes can be easily implemented as a parallel system. Possibility of high-level parallelization can assure another major efficiency improvement.

We investigated MO-mHGS combined with several state-of-the-art MOEAs, including group of standard, generation-based algorithms applying Pareto ranking schemes and few more experimental approaches. One of them was representative of the Particle Swarm Optimization (PSO) family, OMOPSO [12]. Its perfor-mance was not satisfying in comparison to some other considered MOEAs, but after applying it as MO-mHGS engine we received very good outcomes. In final simulation, it gained top scores, regardless of quality metric. PSO algorithms were never before examined in context of hierarchical multi-deme strategies.

These promising results clearly determine a direction of further research. We would like to extend MO-mHGS in order take advantage of its properties we discussed and improve its performance and adaptability to practical problems. We will test following hypotheses:

1. MO-mHGS can be applied as a generic tool for handling decision-making Multi-objective Optimization Problems — this statement is based on our previous experiments conducted during preparation of the research pa-per [7].

2. Particle Swarm Optimization algorithms combined with MO-mHGS out-perform other classes of MOEAs — this statement is also based on our up-to-date conducted experiments. Being aware of Wolpert and MacReady so called No free lunch theorem [15] we will focus on performing broadly planned experiments using acclaimed benchmarks in order to make sure that the proposed computing paradigm prevails in a significant number of tests.

3. If we integrate and optimize hierarchic strategy with PSO solution it will improve results in reference to generic MO-mHGS with PSO engine — this statement is a consequence of our previous research and we are convinced that further work on hybridization of PSO with HGS-related computing system will result in achieving good results (again we will remember about No Free Lunch Theorem while conducting this research).

4. Preparing parallel implementation of the proposed model would significantly increase efficiency of the solution and improve simulation process. The statement is based on characteristics of hierarchical architecture and multi-deme model. Each HGS node manages its own population and conducts evolution process in isolation from other demes. Autonomy of tree components can be used in distributed environment allowing to perform calculations independently, at the same time.

5. Management of population of MO-mHGS can be enhanced towards autonomy, constituting agent-based computing system (similar to EMAS [13] paradigm), minimizing (or even removing) the global control mechanism, yielding a new, efficient agent-based computing system, and preparing it for further testing, also versus the already present MO EMAS implementation. The statement is based on the already conducted research on EMAS paradigm, and apparent efficiency of the system is expected thanks to lower-ing of the global control over the parts of the algorithm. Another important issue of such computing system would be broader exploration and better exploitation of particular, interesting parts of the search space (thanks to HGS features).

In order to meet our objectives, we will extend current research environment and add several new algorithms from PSO family to our test methods set. After preliminary research we will focus on creating new model, MO-HoPSO integrating MO-mHGS with PSO solution using specifics of both strategies.

# 2
# MOEA simulation tools

Developing and evaluating new MOEA solutions requires reliable simulation environment. Such environment should provide large set of tools utilized in the MOEA research methodology. We can distinguish several types of a typical MOEA platform's elements:

- Existing MOEA algorithms implementations - they predominantly play a key role when choosing the right platform for evaluation. Number of ready to use MOEAs may not be very high, though the range of represented algorithm classes should be wide enough to include either classic and current state of the art solutions. Preferably, algorithms should be decomposed into reusable elements such as mutation/crossover operators and selection strategies.

- Benchmarks - multi-objective problems definitions with known Pareto sets. They may vary in number of objectives, but also be constructed in a way to detect vulnerabilities of investigated strategy, e.g. problems with multi-modal Pareto fronts.

- Quality indicators - metrics that can be calculated over obtained results, evaluating their various aspects. It is important to update quality indicators base, as state of the art metrics tend to change.

- Simulation execution tools - it is important to control simulation process, by e. g. managing outcomes serialization and ensuring parallelization sup-port where it is possible. If MOEA library offers such tools it becomes a simulation platform.

- Result processing tools - this element may not be considered obligatory, as it is hard to generalize outcomes interpretation and further processing. Nevertheless, the ability to quickly produce plots of quality indicators or results comparison summary can be very helpful. Especially if we consider many-objectives problems where we have to face high number of output vectors' dimensions and it is necessary to apply more sophisticated results visualization methods.

There is also an another important, non-functional aspect of such platforms that should be considered: accessibility. On the one hand, we would like to operate on large set of existing elements, combining them and extending their functional-ity. On the other, it should be as simple as possible when we want to just prepare simulation run. Additionally, solutions written in high-level languages like Matlab or Python have naturally more intuitive (declarative) syntax than low-level C or even certain Java implementations.

Throughout the years of MOEA domain development, several MOEA plat-forms have been proposed and gained popularity. In jMetal there is a large library of evolutionary algorithms (with significant number of MOEA representatives), implemented in Java. Some of them have built-in parallelization support, but there are no dedicated simulation tools. MOEA Framework provides such mech-anism, as well as flexible extendible programming interface, decomposed to many reusable components. The platform has great support and is still developed. Un-fortunately, for now, it does not provide parallelization API. PlatEMO is a re-cently released library of open-source, Matlab implementation of over 50 state of the art MOEAs and MOPs. It offers also basic visualization tools, but there is no advanced simulation support. Fast implementations of MOEAs written in C are included in OTL (though the authors recommend using Python bindings).

None of aforementioned solutions provides generic meta-models hybridiza-tion mechanism. In MOEA Framework there is possibility to construct an island model, because populations can be modifiable, but majority of MOEA Frame-work code is not thread-safe and it requires additional precautions. ECJ library offers asynchronous island models over TCP/IP, but it is more concentrated on single-objective problems evaluation, providing only classic MOEAs (NSGA-II, SPEA2).

# 3

# Flexible MOEA hybridization model

Working with complex, high-level MOEA meta-models such as Multiobjective Optimization Hierarchic Genetic Strategy (MO-mHGS) [7] with multi-deme support usually requires dedicated implementation and configuration for each internal (single-deme) algorithm variant. If we generalize meta-model, we can simplify whole simulation process and bind any internal algorithm (we denote it as a driver), without providing redundant meta-model implementations. This idea has become a fundamental of Evogil platform. Our aim was to allow constructing custom hybrid models or combine existing solutions in runtime simulation environment. We define hybrid solution as a composition of a meta-model and a driver (or multiple drivers). Meta-model uses drivers to perform evolutionary calculations and process their results (Fig 3.1).

Generalization usually leads to limitations and constraints that reduce its practicality. Therefore, we need to give a possibility to handle characteristics of specific solutions and establish some assumptions about their similarities and requirements of our model. We can expect each MOEA has its own set of configuration parameters. Additionally, multi-deme models can for example take advantage of mixing their populations during simulation process. In other words, we can't assume full separation of drivers and have to handle requests from the outside.

Considering these problems, we can form following hybridization assumptions:

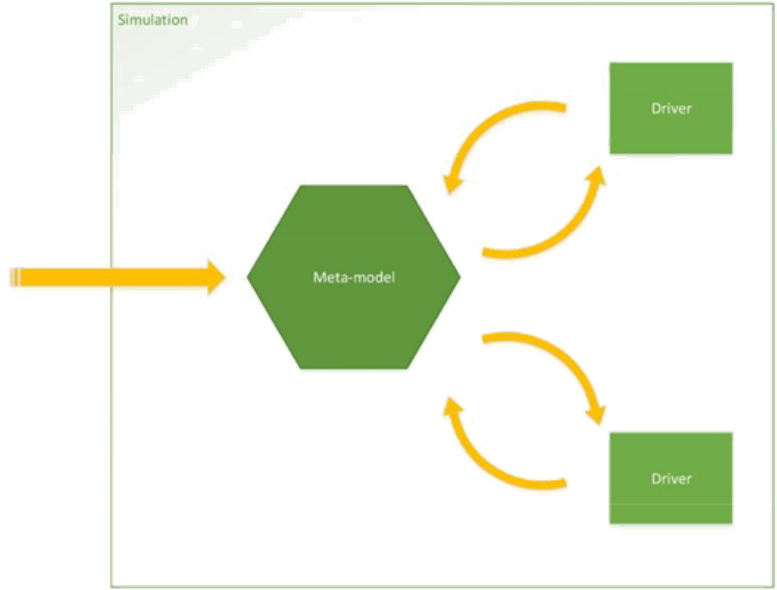

**Figure 3.1:** Evogil hybridization model concept

- Each driver performs its algorithm in repeatable steps. They may be considered as generations or epochs, but steady-state approach [2] can be also adjusted according to one's requirements.

- Driver operates on a population with initial size set by the meta-model. Output population size can be changed during the simulation, but driver should share its current population with the meta-model after each step.

- Meta-model may require additional operations or data from its driver after each step, but it does not influence driver's step algorithm. Therefore, core driver solution is separated from meta-model and can be used in various hybrid combinations.

Note that if meta-model also meets all these requirements, it can be applied as a driver for another meta-model. With Evogil, we can compose such multi-level models without complicated and redundant implementation.

On the other hand, if meta-model works on multiple populations, it may be configured to run different driver on each of them. It can also change driver type depending on current simulation stage or outcomes characteristics.

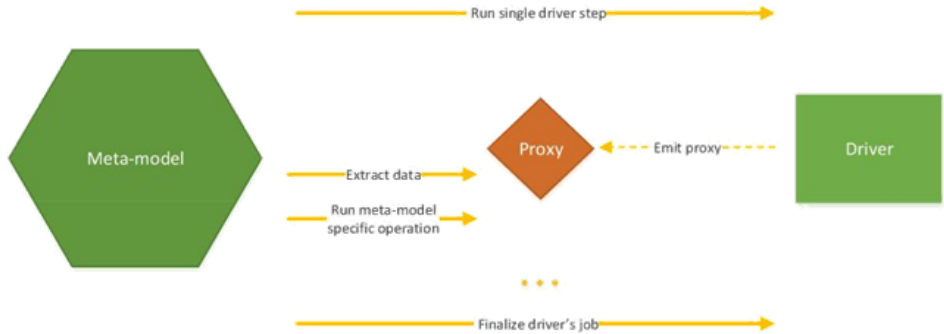

**Figure 3.2:** Evogil hybridization flow. Proxy objects are emitted after each driver's step. Meta-model can invoke specific operations on proxy that has access to required driver's data.

Evogil's hybridization model is built upon flexible message-passing paradigm [?]. Drivers communicate with their superior meta-models by sending proxy objects. A proxy object contains whole data and operations required by the meta-model. After each simulation step, information is exchanged between model levels (Fig 3.2). While the meta-model is working with the received proxy, the driver is awaiting. Then, the meta-model can invoke another step or finalize the driver's job.

In the simplest case, proxy would contain new generation of population developed by the driver in current epoch and additional, statistical information (e.g. epoch cost expressed as number of Fitness function calls).

However, if we considered more sophisticated meta-model such as Island Genetic Algorithm (IMGA) [?], simple approach described above would not be sufficient. In IMGA, driver epochs are interleaved with the migration process. During the migration, individuals are selected from each island (separate driver's instance) and transfered between islands according to the topology configuration. Although migrants selection is managed by IMGA meta-model, it is a driver's responsibility to handle population modification (removing outgoing individuals and assimilate ingoing). Instead of redesigning all drivers that are supposed to work with IMGA, we can place migration-related operations in IMGA-Proxy specific for given driver type. In worst case scenario, we would have to deliver different IMGA-Proxy for each driver algorithm, but we still keep drivers' models separated from meta-models (Fig. 3.3).

As another example of Evogil's hybridization we will consider MO-mHGS model. Unlike IMGA, it does not keep fixed number of drivers developing their populations through the whole simulation. Instead, it dynamically creates new driver nodes in tree-like structure, increasing search precision with every new tree level. New nodes (so called sprouts) manage populations built around "the most promising" individuals from current-level driver's population. In this case driver's population is not modified, but nominating the best individuals is delegated to the driver. Again, this behavior can be achieved by preparing HGS-Proxy object that utilizes driver-specific methods in order to extract individuals and pass them into MO-mHGS for further sprouting processing (Fig. 3.4). MO-mHGS dynamic nature does not collide with Evogil's assumptions because it runs all drivers in steps (meta-epochs) and whole process in repetitive.

In general, introducing proxy objects as communication layer between drivers and meta-models allows to keep algorithms separated from each other. Driver does not posses any knowledge about meta-model (does not even know it is a part of more complex solution) and meta-model operates only on driver's proxies without additional assumptions about driver's type.

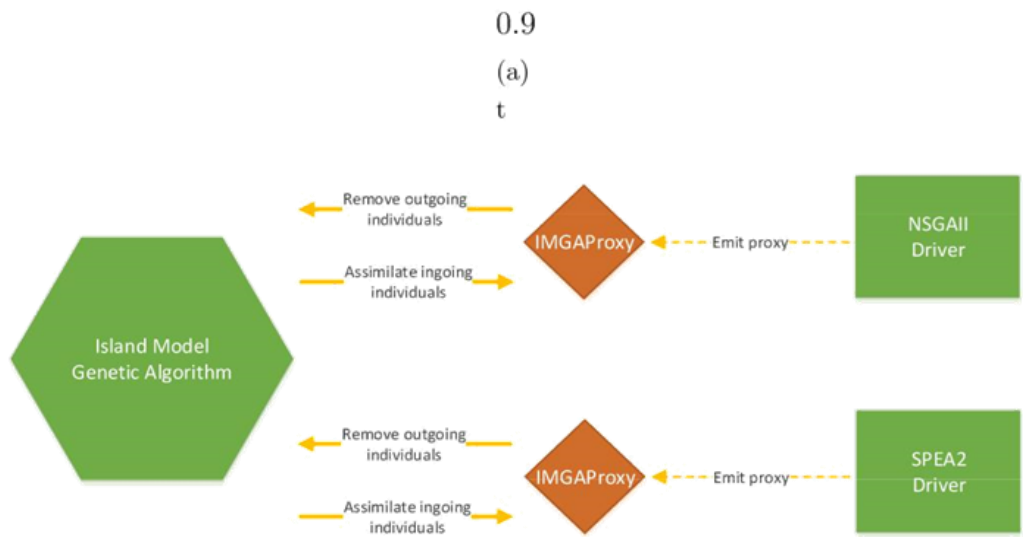

Figure 3.3: Island Model Genetic Algorithm with NSGAII and SPEA2 drivers.

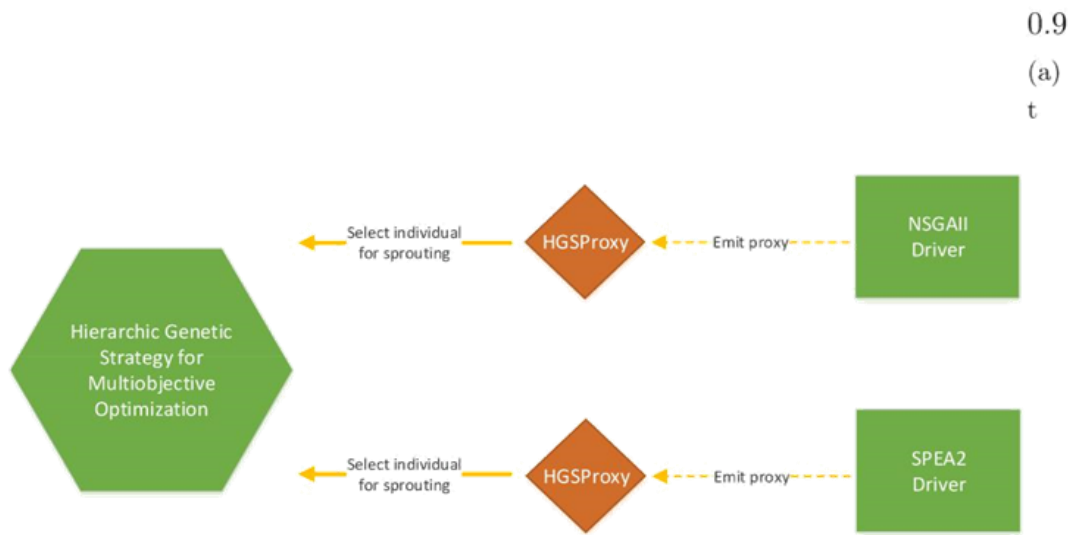

Figure 3.4: Hierarchic Genetic Strategy for Multiobjective Optimization with NSGAII and SPEA2 drivers.

Figure 3.5: Example applications of Evogil hybridization models.

# 4

# Evogil platform evaluation

## 4.1 Simulation components description

Evogil was designed not only another library of evolutionary tools, but more as a framework one can extend and use for his own purposes. It provides set of ready-made solutions divided into two groups (multi-deme meta-models and single-deme drivers), as well as processing tools (quality metrics, statistics and plotting scripts), simulation management and results persistence layer.

Currently it contains implementations of well known algorithms representing different groups of MOEA solutions:

- Non-dominated Sorting Genetic Algorithm (NSGA-II and θ-NSGA-III)

- Strength Pareto Evolutionary Algorithm 2 (SPEA2)

- Multi-objective PSO (OMOPSO)

- Indicator-Based Evolutionary Algorithm (IBEA)

- Smetric selection EMOA (SMS-EMOA)

- Approach based on non-dominated sorting and local search (NSLS)

To take advantage of proposed hybridization model, three different multi-deme meta-models were also provided:

- Multiobjective Optimization Hierarchic Genetic Strategy (MO-mHGS)

- Island Model Genetic Algorithm (IMGA)

- Jumping Gene Based Learning (JGBL)

All mentioned solutions were implemented in Python as a part of Evogil plat-form and thus they meet hybridization requirements we discussed. Any of them can be treated as a driver for any of multi-deme meta-models. Moreover, Evogil's multi-models are also full-fledged drivers so multi-layered composition is possible.

In order to perform simulations, additional components are required. Evogil provides set of testing data, including ZDT benchmarks [17] and DTLZ bench-marks (CEC09 algorithm contest test problems) [16].

Simulation results are gathered incrementally and stored in small chunks. By default Python .pickle file format is used in order to achieve the best performance. Nevertheless, it is also possible to easily change output format to e. g. JSON. Keeping results as independent small portions of information reduces problems with running Evogil in distributed environment. Results can be processed and merged at any time in any amount.

Results processing involves the use of various quality metrics: Generational Distance (GD) [14], Inverted Generational Distance (IGD) [1], Average Haus-dorff Distance (AHD) [11], Hypervolume (HV) [3], Pareto Dominance Indicator (PDI) [6] and Spacing [10]. Metrics values can be also cached by updating out-comes files.

Finally, the output can be visualized using Evogil's plotting tools, allowing to create algorithms outcomes comparison in time for given benchmark, comparison summary for all benchmarks, results distribution in relation to Pareto front or violin plots showing error distribution.

## 4.2 Evogil workflow

We can work with Evogil in two ways: by running existing tools and solutions or by extending the platform with new algorithms and features. The latter involves more technical details and is out of scope of this paper. Therefore, we will focus on understanding Evogil's workflow and its advantages.

Figure 4.1 shows typical Evogil's simulation process. At the beginning, we need to specify algorithms configuration. Usually MOEAs have several parameters we can control. In Evogil they can be specified in simulation configuration file. What is important, we can set different sets of parameters for different hybridization combinations. In example shown in Figure 4.2 there are three sets of parameters: one for bare NSGAII, one for HGS+NSGAII and one for JGBL. In such case JGBL combined with any algorithm will run with the same configuration. We could have also specified specific sets of algorithms' parameters for particular benchmark problem.

Simulations in Evogil can be run using simple command line tool. We can specify i. e. algorithms and hybrid models, benchmarks and budget. We will discuss simulation budgets later, but first we consider simplest working example:

```
evogil.py    500 -a   NSGAII -p ZDT1
```

It starts simulation of NSGAII algorithm on ZDT1 benchmark width budget equal to 500. If instead we would like to run simulation with hybrid model, using NSGAII as a driver of MO-mHGS, we would use following command:

```
evogil.py   500 -a HGS+NSGAII -p ZDT1
```

Hybrid models can be multi-layered:

```
evogil.py 500 -a JGBL+HGS+NSGAII -p ZDT1
```

In such case, HGS+NSGAII is treated as a driver for JGBL. According to the configuration we specified in Fig. 4.2, JGBL will use generic set of parameters and HGS+NSGAII will obtain specific configuration (therefore, bare NSGAII configuration will not be used here).

Simulation bounds are determined by its budget. Budget specifies simulation duration in units that do not need to be related to actual elapsed time. Budget is forwarded to the algorithm where it should be consumed according to adopted strategy. If budget is depleted, simulation ends.

By default, budget is consumed by each fitness function call - in this way we can observe which solution obtained the best results minimizing number of fitness calculations (that can be expensive in real-world). However, we can adopt simpler approach that consumes 1 unit of budget after each algorithm step. In this case, budget value is equal to e. g. number of epochs.

We can also specify several budget checkpoints. After reaching a checkpoint, simulation creates a snapshot of most recent algorithm's results and stores them in persistence layer. It is very useful for further results analysis, especially generating plots visualizing metrics values over specific budget constraints. In our example, we would extend evogil command as follows:

```
evogil.py    100,300,500 -a JGBL+HGS+NSGAII -p ZDT1
```

As a result, Evogil would generate 3 result checkpoint files for budget values 100, 300 and 500.

Apart from splitting budget for single simulation, it is also possible to run multiple solutions for multiple problems at once:

```
evogil.py              100,300,500 -a  NSGAII,
JGBL+HGS+NSGAII -p ZDT1, UF1
```

Each simulation is a triple (budgets, algorithm, problem) so in above example 4 simulations will be run. Naturally, initial budget values are copied to each simulation separately. We can also set -N X parameter to repeat each simulation X times.

By default Evogil is managed sequentially, but by adding flag -j Y to the above example we can create Y processes pool unleashing Evogil parallelization capabilities.

While Evogil simulation produces outcomes, it is possible to analyze them in another process, adding quality metrics, etc. When the simulation is finished, result files can be gathered and processed by other Evogil modules (such as Plotter or Summary Generator) or external applications.

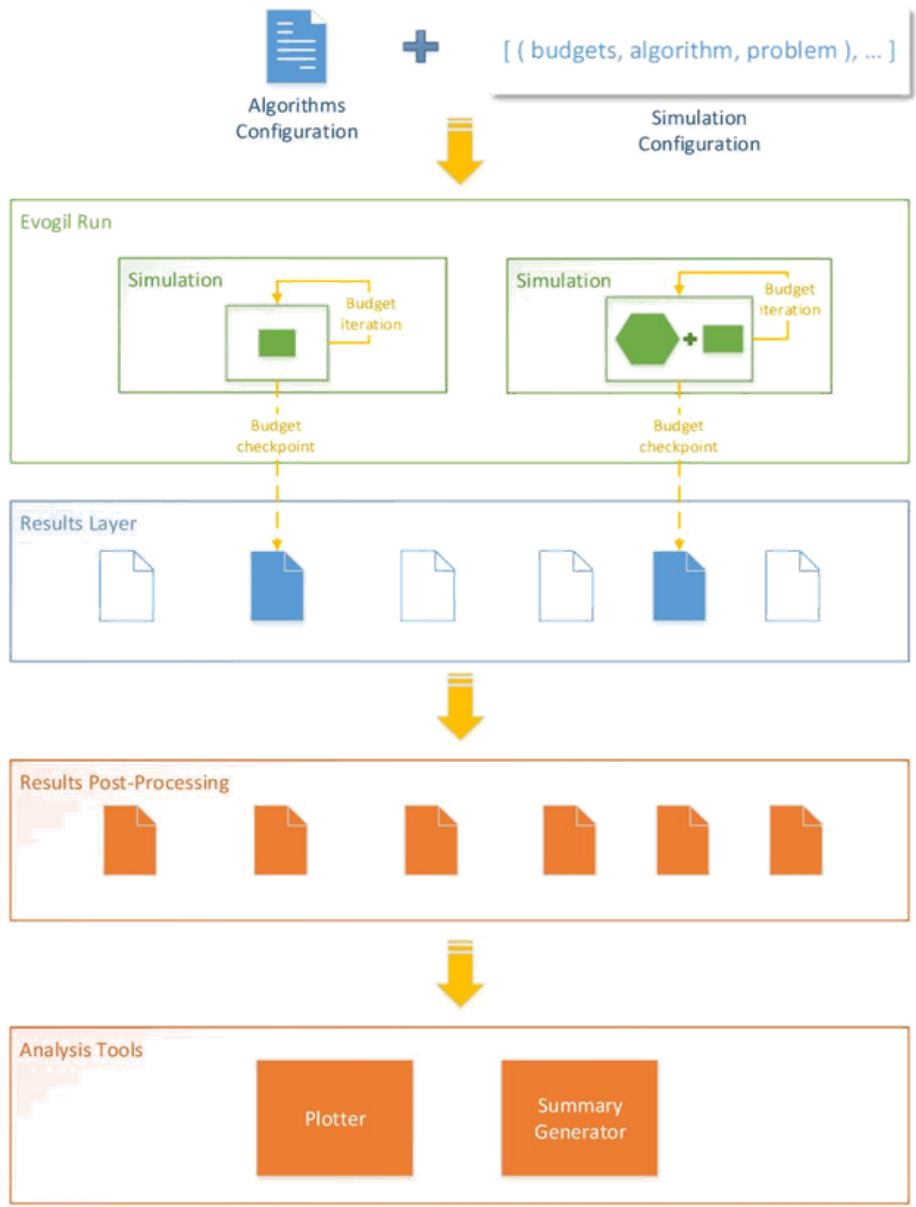

Figure 4.1: Evogil simulation workflow.

```
"NSGAII": {
  "mating_population_size": 0.5
},

"JGBL": {
  "mating_population_size": 0.5,
  "jumping_rate": 0.6,
  "jumping_percentage": 0.5
},

"HGS + NSGAII": {
  "mating_population_size": 0.4,
  "max_sprouts_no": 16,
  "sproutiveness": 3,
  "metaepoch_len": 5
}
```

FIGURE 4.2: Evogil's simulation example configuration containing simulation parameters for bare NSGAII algorithm, HGS+NSGII hybrid model and all JGBL hybrid models.

# 5
# Evogil contents

Below you can find full listing of Evogil platform's contents. The platform is still under development and will be extended with new elements in the future. Moreover, the project is open–source and any person can contribute, adding additional implementations of algorithms or real-world multi-objective problems.

## 5.1   Algorithms

- Non-dominated Sorting Genetic Algorithm (NSGA-II and θ-NSGA-III)
- Strength Pareto Evolutionary Algorithm 2 (SPEA2)
- Multi-objective PSO (OMOPSO)
- Indicator-Based Evolutionary Algorithm (IBEA)
- Smetric selection EMOA (SMS-EMOA)
- Aproach based on nondominated sorting and local search (NSLS)

## 5.2   Multi-deme meta-models

- Hierarchical Genetic Strategy (HGS)
- Distributed Hierarchical Genetic Strategy (DHGS)

- Island Model Genetic Algorithm (IMGA)
- Jumping Gene Based Learning (JGBL)

## 5.3 Quality indicators

- Hypervolume
- Generational Distance (GD)
- Inverse Generational Distance (IGD)
- Average Hausdorff Distance (AHD)
- Epsilon
- Extent
- Spacing
- Pareto Dominance Indicator (PDI)

## 5.4 Benchmarks

- ZDT family (ZDT1, ZDT2, ZDT3, ZDT4, ZDT6)
- cec2009 family (UF1-UF12)
- kursawe
- ackley

# 6
# Conclusion

Evogil platform was designed in order to combine popular MOEAs with complex, multi-deme models such as MO-mHGS or island-based model. It can be also applied as a main tool for managing parallel simulation process or gathering and analyzing results. We plan to use Evogil components in this project. We will also extend the platform to include missing functions and components. We plan to use them in this project and also extend Evogil platform to include missing functions and components. It is also worth to note that Evogil platform provid-ing technical support for our research has been already shared as an open-source project at Github. It assures reproducibility of the results we describe and opens up possibility of extending the project or its application in other research tasks.